\def\year{2018}\relax
\documentclass[letterpaper]{article} 
\usepackage{aaai18}  
\usepackage{times}  
\usepackage{helvet}  
\usepackage{courier}  
\usepackage{amssymb}
\usepackage{graphicx}
\usepackage{amsmath}
\usepackage{algorithm}
\usepackage{algorithmic}
\usepackage{mathtools}
\usepackage{stmaryrd}
\usepackage{subcaption}
\usepackage{adjustbox}
\usepackage{multirow}
\usepackage{enumitem}
\usepackage{booktabs} 
\usepackage{url}
\usepackage{amsthm}
\usepackage{balance}
\usepackage{flushend}
\newtheorem{definition}{Definition}

\newcommand{\ours}[0]{$\text{M2E}$}
\providecommand{\keywords}[1]{\textbf{\textit{Index terms---}} #1}

\begin{document}

\title{Multi-View Multi-Graph Embedding for Brain Network Clustering Analysis}

\author{Ye Liu$^1$, Lifang He$^{2}\thanks{Corresponding Author}$, Bokai Cao$^1$, Philip S. Yu$^{1,3}$, Ann B. Ragin $^4$, Alex D. Leow$^5$\\
$^1$Department of Computer Science, $^5$Department of Bioengineering, University of Illinois at Chicago, IL, USA\\
$^2$Department of Healthcare Policy and Research, Cornell University, NY, USA\\
$^3$Institute for Data Science, Tsinghua University, Beijing, China\\
$^4$Department of Radiology, Northwestern University, IL, USA\\
\{yliu279, caobokai, psyu\}@uic.edu, \{lifanghescut, alexfeuillet\}@gmail.com, ann-ragin@northwestern.edu 
}

\maketitle

\begin{abstract}
Network analysis of human brain connectivity is critically important for understanding brain function and disease states. Embedding a brain network as a whole graph instance into a meaningful low-dimensional representation can be used to investigate disease mechanisms and inform therapeutic interventions. Moreover, by exploiting information from multiple neuroimaging modalities or views, we are able to obtain an embedding that is more useful than the embedding learned from an individual view. Therefore, multi-view multi-graph embedding becomes a crucial task. Currently only a few studies have been devoted to this topic, and most of them focus on vector-based strategy which will cause structural information contained in the original graphs lost. As a novel attempt to tackle this problem, we propose Multi-view Multi-graph Embedding ({\ours}) by stacking multi-graphs into multiple partially-symmetric tensors and using tensor techniques to simultaneously leverage the dependencies and correlations among multi-view and multi-graph brain networks. Extensive experiments on real HIV and bipolar disorder brain network datasets demonstrate the superior performance of {\ours} on clustering brain networks by leveraging the multi-view multi-graph interactions. 
 
\end{abstract}
\keywords{Brain Network Embedding, Multi-graph Embedding, Tensor Factorization, Multi-view Learning}

\section{Introduction}
Benefiting from modern neuroimaging technology, there is an increasing amount of graph data representing the human brain, called brain networks, \emph{e.g.}, functional magnetic resonance imaging (fMRI) and diffusion tensor imaging (DTI). These data have complex structure, which are inherently represented as graphs with a set of nodes and links. Moreover, the linkage structure extracted from different modalities can often be treated as multi-view data. The connections in fMRI brain networks encode correlations among brain regions in terms of functional activities, while in DTI networks, the connections can capture the white matter fiber pathways that connect different brain regions. Even if these individual views might be sufficient on their own for a given learning task, they can often provide complementary information to each other which can lead to improve performance on the learning task \cite{ma2017multi,sun2017sequential}. As labeled data are difficult to obtain, it is critical to leverage the multi-view information to obtain an effective embedding for the clustering task. Therefore, in this study, we focus on investigating the multi-view multi-graph embedding problem for brain network clustering analysis. Specifically, we aim to learn the latent embedding representation of multiple brain networks from multiple views.
In recent years, there has been an increasing interest in single-graph node embedding among researchers \cite{mousazadeh2015embedding,ou2016asymmetric}, yet it is challenging to extend to the multi-graph embedding, where we consider the embedding of multiple graph instances together to obtain a discriminative representation for each graph. 
By exploring the consistency and complementary properties of different views, multi-view learning \cite{liu2013multi,cao2014tensor} is rendered more effective, more promising, and has better generalization ability than single-view learning. Although there have been numerous works on single-graph embedding and multi-view learning, to the best of our knowledge, there is no embedding method available which enables preserving multi-graph structures on multiple views.

There are several major challenges in multi-view multi-graph embedding problem. Initially, the complex graph structure makes conventional methods difficult to capture the subtle local topological information \cite{jie2014integration}. For the subgraph based method \cite{cao2015identifying}, the number of subgraphs is exponential to the size of the graphs. Thus the subgraph enumeration process is both time and memory consuming. Besides, simply preserving pairwise distances, as with many spectral methods, is insufficient for capturing the structure of multiple graphs. Moreover, preserving both local distances and graph topology is crucial for producing effective low-dimensional representations of the brain network data. Furthermore, traditional normalization strategies cannot generate meaningful clustering results. 

\begin{figure}[t]
\centering
\includegraphics[width=9cm]{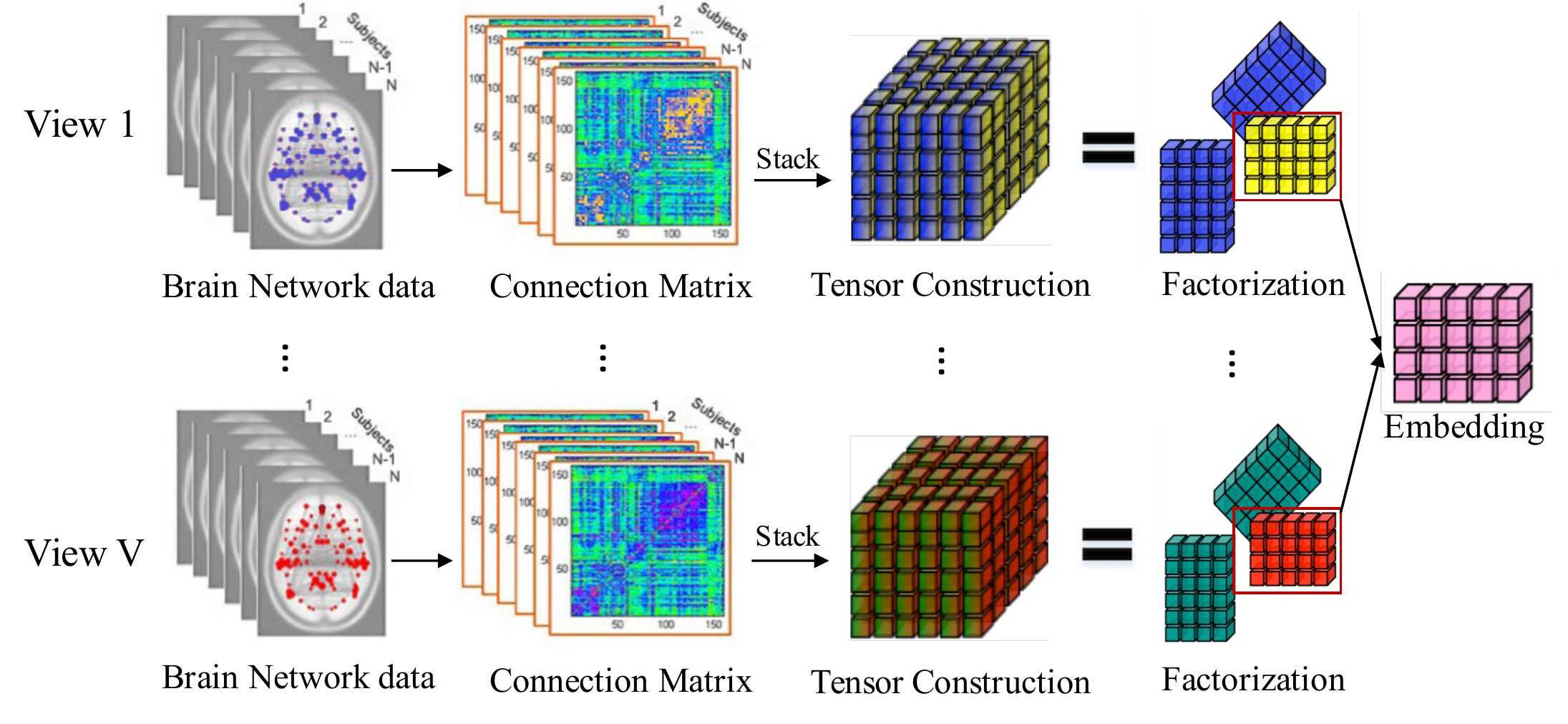}
\caption{A conceptual view of Multi-view Multi-graph Embedding (M2E)}
\label{fig:Frame}
\end{figure}

To address the aforementioned issues, in this paper we propose a novel one-step Multi-view Multi-graph Embedding ({\ours}) approach for brain network analysis. The goal of {\ours} is to find low-dimensional representations from multi-view multi-graph data which reveal patterns and structures among the brain networks. The conceptual view of {\ours} is shown in Figure \ref{fig:Frame}. Our main contributions are summarized as follows:

\begin{itemize}[leftmargin=*]
    \item In each view, we stack all the brain networks on each view into a tensor and use tensor and matrix techniques to simultaneously leverage the dependencies and correlations among multi-view and multi-graph data in a unified framework. This provides an innovative perspective on the analysis of brain network structures.
    \item In order to reflect the latent clustering structure shared by different views and different graphs, we require coefficient matrices learned from different views towards a consensus with soft regularization.
    \item We present an effective optimization strategy to solve the {\ours} problem, with consideration of symmetric structure of brain networks.
\end{itemize}
Through extensive experiments on HIV and bipolar disorder brain network datasets that contain fMRI and DTI views,
we demonstrate that {\ours} can significantly boost the embedding performances. Furthermore, the derived factors are visualized which could be informative for investigating disease mechanisms.

\section{Related Work}\label{sec:related}
Brain Network Embedding makes characterization of brain disorders at a whole-brain connectivity level possible, thus providing a new direction for brain disease clustering. The goal of graph embedding is to find low-dimensional representations of nodes that can preserve the important structure and properties of graphs \cite{ma2016multi}. In particular, \cite{mousazadeh2015embedding} proposed a graph embedding algorithm based on Laplacian-type operator on manifold, which can apply to recover the geometry of data and extend a function on new data points. Recently, \cite{ou2016asymmetric} established a general formulation of high-order proximity measurements, and then applied it with generalized SVD for graph embedding. In the field of brain network neuroscience, most of the existing works aim to learn the structure from a specific kind of brain networks \cite{kong2014brain,kuo2015unified}. In contrast to the node embedding on single graph, we aim at learning an effective graph embedding approach on multi-view multi-graph brain networks, such as fMRI brain network together with DTI brain network.

Our approach is also closely related to the literature on multi-view clustering and multi-view embedding. \cite{kumar2011traing,kumar2011co} are the first works proposed to solve the multi-view clustering problem via spectral projection. \cite{AMGL} extended the multi-view spectral clustering to a parameter-free auto-weighted method. Matrix factorization based methods \cite{liu2013multi} are another category, which mainly use non-negative matrix factorization (NMF) to integrate multi-view data. Additionally, \cite{shao2015clustering} proposed a CP factorization and $\ell_1$-norm regularization based method for multi-view incomplete clustering. \cite{ma2017multi2} coupled the spectral clustering and $\ell_{2,1}$-norm to discriminate the hubs and to reduce the potential influence of the hubs for graph clustering. However, there is no method available which enables us to take multiple graphs as input and consider multi-graph structures; thus multi-view learning cannot solve the brain network embedding problem well. 

\section{Preliminaries} \label{sec:pre}
In this section, we introduce some related concepts and notation about tensor. Table \ref{tab:notation} lists basic symbols that will be used throughout the paper.

Tensors are higher order arrays that generalize the notion of vectors and matrices. The order of a tensor is the number of dimensions (a.k.a. modes or ways).
An $M$-th order tensor is represented as $\mathcal{X} \in \mathbb{R}^{I_{1} \times \cdots \times I_{M}}$, where $I_m$ is the cardinality of its $m$-th mode, $m \in [1:M]$. All vectors are column vectors unless otherwise specified. For an arbitrary matrix $\mathbf{X} \in \mathbb{R}^{I \times J}$, its $i$-th row and $j$-th column vector are denoted by $\mathbf{x}^{i}$ and $\mathbf{x}_{j}$, respectively.

Definitions of partially symmetric tensor, mode-$m$ matricization and CP factorization are given below, which will be used to present our model.

\begin{definition}[Partial Symmetric Tensor]
An $M$-th order tensor is a rank-one partial symmetric tensor if it is partial symmetric on modes $i_1,...,i_j \in [1:M]$, and can be written as the tensor product of $M$ vectors, \emph{i.e.},
\begin{align}
\mathcal{X}=\mathbf{x}^{(1)} \circ \cdots \circ \mathbf{x}^{(M)}
\end{align}
where $\mathbf{x}^{(i_1)}=\cdots=\mathbf{x}^{(i_j)}$.
\end{definition}

\begin{definition}[Mode-$m$ Matricization] The mode-$m$ matricization of a tensor $\mathcal{X} \in \mathbb{R}^{I_{1} \times \cdots \times I_{M}}$, denoted by $\mathbf{X}_{(m)} \in \mathbb{R}^{I_m \times J}$, where $J = \Pi_{q=1, q \neq m}^M I_q$. Each tensor element with indices ($i_1, \cdots, i_M$) maps to a matrix element ($i_m, j$), such that
\begin{align}
j & = 1+\sum_{p=1,p\neq m}^M(i_p-1)J_p, ~with \nonumber\\
J_p & = \begin{cases} 
  1,  & \mbox{if } p =1 ~\mbox{or}~(p=2 ~\mbox{and}~m = 1)  \\
  \Pi_{q=1, q \neq m}^{p-1}I_q, & \mbox{otherwise}. 
\end{cases}
\end{align}
\nonumber
\end{definition}

\begin{table}[t]
\centering
\caption{List of basic symbols.}
 \resizebox{3.5in}{!}{
\label{tab:notation}
\begin{tabular}{ll}
\toprule
Symbol & Definition and Description\\
\midrule
$x$ & each lowercase letter represents a scale\\
$\mathbf{x}$ & each boldface lowercase letter represents a vector\\
$\mathbf{X}$ & each boldface uppercase letter represents a matrix\\
$\mathcal{X}$ & each calligraphic letter represents a tensor, set or space\\
$[1:M]$ & a set of integers in the range of $1$ to $M$ inclusively. \\
$\circ$ & denotes the outer product \\
$\odot$ & denotes Khatri-Rao product\\
$\llbracket \cdot \rrbracket$ & denotes the CP factorization \\
\bottomrule
\end{tabular}}
\end{table}

\begin{definition}[CP Factorization]
For a general tensor $\mathcal{X}\in \mathbb{R}^{I_1\times\cdots\times I_M}$, its CANDECOMP/PARAFAC (CP) factorization is 
\begin{align}
\mathcal{X}  \approx   \sum_{r=1}^R \mathbf{x}_{r}^{(1)} \circ \cdots \circ \mathbf{x}_{r}^{(M)} \equiv \llbracket \mathbf{X}^{(1)},...,\mathbf{X}^{(M)} \rrbracket
\end{align}
where for $m\in [1:M]$, $\mathbf{X}^{(m)}=[\mathbf{x}_1^{(m)},...,\mathbf{x}_R^{(m)}]$ are latent factor matrices of size $I_m \times R$, $R$ is the number of factors, and $\llbracket \cdot \rrbracket$ is used for shorthand.
\end{definition}

To obtain the CP factorization $\llbracket \mathbf{X}^{(1)}, \cdots, \mathbf{X}^{(M)} \rrbracket$, the objective is to minimize the following estimation error:
\begin{align}
     \mathcal{L} = \underset{\mathbf{X}^{(1)}, \cdots, \mathbf{X}^{(M)}}{\min} \| \mathcal{X} - \llbracket \mathbf{X}^{(1)}, \cdots, \mathbf{X}^{(M)} \rrbracket \|_F^2 \label{eq:CP_problem}
\end{align}
However, $\mathcal{L}$ is not jointly convex w.r.t. $\mathbf{X}^{(1)}, \cdots, \mathbf{X}^{(M)}$. A widely used optimization technique is the Alternating Least Squares (ALS) algorithm, which alternatively minimize $\mathcal{L}$ for each variable while fixing the other, that is,
\begin{equation}
\mathbf{X}^{(k)} \leftarrow \underset{\mathbf{X}^{(k)}}{\arg\min} \| \mathbf{X}_{(k)} - \mathbf{X}^{(k)} (\odot_{i \neq k}^n \mathbf{X}^{(i)} )^\mathrm{T} \|_F^2 \label{eq:ALS}
\end{equation}
where $\odot_{i \neq k}^M \mathbf{X}^{(i)} = \mathbf{X}^{(M)} \odot \cdots \mathbf{X}^{(k-1)} \odot \mathbf{X}^{(k+1)} \cdots \odot \mathbf{X}^{(1)}$.

\section{Methodology} \label{sec:method}
In this section, we first define the problem of interest. Then we formulate the proposed Multi-view Multi-graph Embedding ({\ours}) method. Finally, we introduce an effective optimization approach to solve the proposed formulation.

\subsection{Problem Definition}
We study the problem of multi-view multi-graph embedding for brain network clustering analysis. Suppose that the problem includes $N$ subjects with $V$ views, where each view has a set of $N$ symmetric brain networks corresponding to $N$ subjects. Specifically, each brain network is represented as a weighted undirected graph, {\em i.e.}, a symmetric affinity matrix $\mathbf{W} \in \mathbb{R}^{M \times M}$ where $M$ denotes the number of nodes and each element reflects connectivity between nodes. There exists a one-to-one mapping between nodes in different graphs, which means that all the graphs have a common node set $M$. Thus, for the $v$-th view, we have $N$ graphs associated with $N$ affinity matrices, denoted as $\mathcal{D}^{(v)}=\{\mathbf{W}_1^{(v)},\mathbf{W}_2^{(v)},\cdots ,\mathbf{W}_N^{(v)}\}$. We use $\mathcal{D}=\{\mathcal{D}^{(1)},\mathcal{D}^{(2)},\cdots, \mathcal{D}^{(V)}\}$ to represent the  multi-view multi-graph instances.

The goal of this work is to learn a common embedding across all brain networks, denoted as $\mathbf{F}^* \in \mathbb{R}^{N \times R}$, where $R$ is the embedding dimension and each row of $\mathbf{F}^*$ corresponds to an embedding of a brain network as a whole. More specifically, we aim at finding $\mathbf{F}^*$ by simultaneously leveraging the dependencies and correlations among multiple views and multiple graphs in $\mathcal{D}$, and taking into account the symmetric property of brain networks. In particular, we investigate the use of learned embedding $\mathbf{F}^*$ for clustering brain networks. Let the number of clusters be $K$. So, we cluster $N$ brain networks into $K$ groups.

\subsection{{\ours} Approach} \label{sec:approach}
Solving challenging multi-view multi-graph embedding problem requires the use of ``complex'' structured models -- those incorporating relationships between multiple views and multiple graphs. The multi-mode structure of tensor provides a natural way to encode the underlying multiple correlations between data \cite{he2017kernelized}. Inspired by the success of tensor analysis on many structured learning problems, here we explore the use of tensor operator techniques to consider all possible dependence relationships among different views and different graphs.

Given a multi-view multi-graph dataset $\mathcal{D}=\{\mathcal{D}^{(1)},\mathcal{D}^{(2)},\cdots, \mathcal{D}^{(V)}\}$. 
In order to capture the multi-graph structures directly, we concatenate the affinity matrices of different subjects for each view to form a third-order tensor comprising three modes: nodes, nodes, and subjects, denoted as $\mathcal{X}^{(v)} = [\mathbf{W}_1^{(v)}, \mathbf{W}_2^{(v)}, \cdots, \mathbf{W}_N^{(v)}]$ $\in \mathbb{R}^{M \times M \times N}, v\in [1:V]$. Notice that since each brain network is a symmetric network, thus the resulting tensor is a partial symmetric tensor. 

Tensor provides a natural and efficient representation for multi-graph data, but there is no guarantee that such representation will be good for subsequent learning, since learning will only be successful if the regularities that underlie the data can be discerned by the model \cite{he2014dusk}. In previous work, it was found that CP factorization is particularly effective to acknowledge the connections and find valuable features among tensor data \cite{van2016structured}. Motivated by these observations, in the following we investigate how to exploit the benefits of CP factorization to find an effective embedding $\mathbf{F}^*$ in the sense of multi-view partial symmetric tensors $\mathcal{X}^{(v)}, v\in [1:V]$.
\begin{figure}[t]
\centering
\includegraphics[width=7cm]{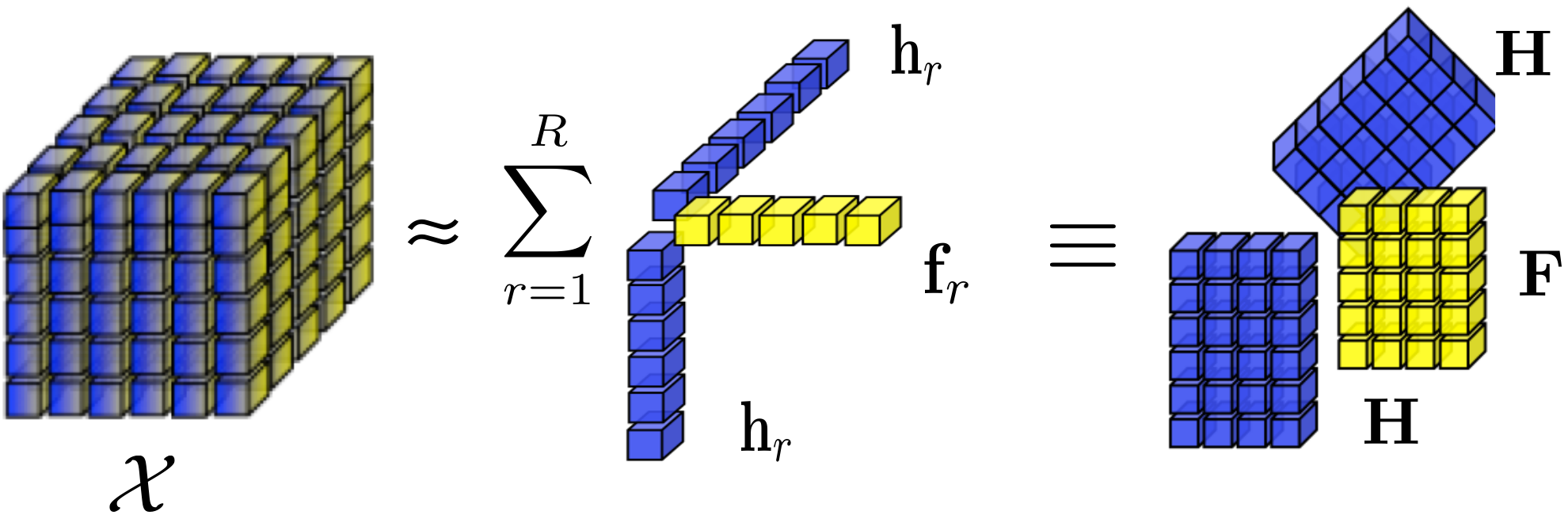}
\caption{CP Factorization. The third-order partially symmetric tensor $\mathcal{X}$ is approximated by $R$ rank-one tensors. The $\emph{r}$-th factor tensor is the tensor product of three vectors, \emph{i.e.}, $\mathbf{h}_r \circ \mathbf{h}_r \circ \mathbf{f}_r$.}
\label{fig:CP}
\end{figure}

A simple method is to learn a view-independent multi-view representation from the tensors $\mathcal{X}^{(v)}$, and then feed it into a conventional multi-view embedding method. This can be formulated as follows: 
\begin{align}
\displaystyle {\min_{\mathbf{H}^{(v)}, \mathbf{F}^{(v)}}} \sum_{v=1}^{V} || \mathcal{X}^{(v)}- \llbracket \mathbf{H}^{(v)}, \mathbf{H}^{(v)}, \mathbf{F}^{(v)} \rrbracket||_F^2
\label{eq:obj1}
\end{align}

where $\mathbf{H}^{(v)} \in \mathbb{R}^{M \times R}$ and $\mathbf{F}^{(v)} \in \mathbb{R}^{N \times R}$ are the latent factor matrices obtained by CP factorization. A graphical representation of this process in one view is given in Figure \ref{fig:CP}. $\mathbf{H}^{(v)}$ can be viewed as common features of nodes among all graphs involved in the $v$-th view, while $\mathbf{F}^{(v)}$ are treated as embedded features of each graph in the $v$-th view.

Based on the above obtained multi-view features $\mathbf{F}^{(v)}$, we can directly establish the following multi-view model to learn a common embedding $\mathbf{F}^*$:
\begin{align}
\displaystyle {\min_{\mathbf{F}^*}} \sum_{v=1}^{V}\lambda_v||\mathbf{F}^{(v)} - \mathbf{F}^*||_F^2
\label{eq:obj2}
\end{align}
where $\lambda_v$ are the weight parameters reflecting the importance of different views.

However, the two-step method, referred to as M2E-TS, is not guaranteed to produce an optimal clustering result, because multiple views and multiple graphs are explored separately. For clustering, we assume that a data point in different views would be assigned to the same cluster with high probability. Therefore, in terms of tensor factorization, we require coefficient matrices learned from different views to be softly regularized towards a common consensus. This consensus matrix is considered to reflect the latent clustering structure shared by different views and different graphs. Based on this idea, we incorporate the Eq. (\ref{eq:obj1}) and Eq. (\ref{eq:obj2}) together and achieve the following optimization problem for {\ours} method:
\begin{align}
\label{eq:obj}
\mathcal{O} = &\displaystyle {\min_{\mathbf{H}^{(v)}, \mathbf{F}^*, \mathbf{F}^{(v)}}} \sum_{v=1}^{V} || \mathcal{X}^{(v)}- \llbracket \mathbf{H}^{(v)}, \mathbf{H}^{(v)}, \mathbf{F}^{(v)} \rrbracket||_F^2 \\ \nonumber
& + \sum_{v=1}^{V}\lambda_v||\mathbf{F}^{(v)} - \mathbf{F}^*||_F^2
\end{align}

Notice that the first term is used to explore the dependencies among multiple graphs, and the second term is used to explore the consensus correlations among multiple views. $\lambda_v$ not only tune the relative weight among different views, but also between the first term and the second term. $\mathbf{F}^*$ is the final embedding solution used for mult-view multi-graph brain network clustering.  To induce groupings on $\mathbf{F}^*$, we simply use $K$-means \cite{hartigan1979algorithm}. 

In order to verify the effectiveness of soft regularization in Eq.~(\ref{eq:obj}), we propose M2E-DS as compared method which learns the latent embedding representation by using the directly shared coefficient matrices $\mathbf{F}^{*}$ for all views \cite{liu2013mining}; the objective function is shown as 
\begin{align}
\displaystyle {\min_{\mathbf{H}^{(v)}, \mathbf{F}^{*}}} \sum_{v=1}^{V} || \mathcal{X}^{(v)}- \llbracket \mathbf{H}^{(v)}, \mathbf{H}^{(v)}, \mathbf{F}^{*} \rrbracket||_F^2
\end{align}
In this formulation, different views are treated equally. However, in reality, different views may have different effects. The detail will be discussed in the Section Experiments.

\subsection{Optimization Framework}
The model parameters in Eq.~(\ref{eq:obj}) that have to be estimated include $\mathbf{H}^{(v)}\in \mathbb{R}^{M \times R}$, $\mathbf{F}^{(v)} \in \mathbb{R}^{N \times R}$ and $\mathbf{F}^* \in \mathbb{R}^{N \times R}$. Since the optimization problem is not convex with respect to $\mathbf{H}^{(v)},\mathbf{F}^{(v)}$ and $\mathbf{F}^*$ together, there is no closed-form solution. We introduce an effective iteration method to solve this problem. The main idea is to decouple the parameters using an Alternating Direction Method of Multipliers (ADMM) approach \cite{boyd2011distributed}. Specifically, the following three steps are repeated until convergence.

\subsubsection{Fixing $\mathbf{F}^{(v)}$ and $\mathbf{F}^*$, compute $\mathbf{H}^{(v)}$}
Note that $\mathcal{X}^{(v)}$ is a partially symmetric tensor and the objective function in Eq.~(\ref{eq:obj}) involving a fourth-order term $\mathbf{H}^{(v)}$ is difficult to optimize directly. To obviate this problem, we use a variable substitution technique and minimize the following objective function
\begin{align}
& \underset{{\mathbf{H}^{(v)}}, \mathbf{P}^{(v)}}{\min} ~|| \mathcal{X}^{(v)}- \llbracket \mathbf{H}^{(v)}, \mathbf{P}^{(v)}, \mathbf{F}^{(v)} \rrbracket ||_F^2 \nonumber \\
&s.t. ~\mathbf{H}^{(v)} = ~\mathbf{P}^{(v)}
\label{HP}
\end{align}
where $\mathbf{P}^{(v)}$ are auxiliary variables.

The augmented Lagrangian function for the problem in Eq.~(\ref{HP}) is 
\begin{align}
& \mathcal{L}(\mathbf{H}^{(v)},\mathbf{P}^{(v)}) = \| \mathcal{X}^{(v)} - \llbracket \mathbf{H}^{(v)}, \mathbf{P}^{(v)}, \mathbf{F}^{(v)} \rrbracket\|_F^2 \nonumber \\
& ~~~~~ + tr(\mathbf{U}^{(v)T} (\mathbf{H}^{(v)} - \mathbf{P}^{(v)}))+\frac{\mu}{2}\|\mathbf{H}^{(v)} -\mathbf{P}^{(v)}\|_F^2
\label{eq:lagrangeHP}
\end{align}
where $\mathbf{U}^{(v)} \in \mathbb{R}^{M \times R}$ are Lagrange multipliers, and $\mu$ is the penalty parameter which can be adjusted efficiently according to \cite{lin2011linearized}.

To compute $\mathbf{H}^{(v)}$, the optimization problem in Eq.~(\ref{eq:lagrangeHP}) can be formulated as
\begin{align}
\underset{\mathbf{H}^{(v)}}{\min}~||\mathbf{X}_{(1)}^{(v)} - \mathbf{H}^{(v)} \mathbf{D}^{(v)\mathrm{T}}||_F^2 + \frac{\mu}{2}||\mathbf{H}^{(v)} - \mathbf{P}^{(v)} +\frac{1}{\mu}\mathbf{U}^{(v)}||_F^2
\label{HE}
\end{align}
where $\mathbf{X}^{(v)}_{(1)} \in \mathbb{R}^{M \times (MN)}$ is the mode-$1$ matricization of $\mathcal{X}^{(v)}$, and $\mathbf{D}^{(v)}=\mathbf{F}^{(v)} \odot \mathbf{P}^{(v)} \in \mathbb{R}^{(NM)\times R}$. 

We rewrite Eq.~(\ref{HE}) in the trace form as
\begin{align}
\underset{\mathbf{H}^{(v)}}{\min}~tr(\mathbf{H}^{(v)} \mathbf{A}^{(v)}  \mathbf{H}^{(v)^\mathrm{T}}) - tr(\mathbf{B}^{(v)^\mathrm{T}} \mathbf{H}^{(v)})
\label{trace}
\end{align}
where $\mathbf{A}^{(v)} = \mathbf{D}^{(v)^\mathrm{T}} \mathbf{D}^{(v)} + \frac{\mu}{2} \mathbf{I}$, and $\mathbf{B}^{(v)} = 2\mathbf{X}^{(v)}_{(1)} \mathbf{D}^{(v)} + \mu \mathbf{P}^{(v)} - \mathbf{U}^{(v)}$.

The problem (\ref{trace}) is a univariate optimization problem, and can be solved easily. An effective approach to solve such a problem is by the proximal gradient method \cite{parikh2014proximal}, which updates $\mathbf{H}^{(v)}$ by
\begin{align}
 \mathbf{H}^{(v)}_{t+1} \leftarrow \mathbf{H}^{(v)}_{t}- \frac{1}{L^{(v)}} (2\mathbf{H}^{(v)^\mathrm{T}}\mathbf{A}^{(v)}-\mathbf{B}^{(v)})
 \label{H}
\end{align}

where $L^{(v)}$ is the Lipschitz coefficient of Eq.(\ref{trace}) that equals to the maximum eigenvalue of $2\mathbf{A}^{(v)}$.

To efficiently compute $\mathbf{D}^{(v)^\mathrm{T}}\mathbf{D}^{(v)}$, we consider the following property of the Khatri-Rao product of two matrices
\begin{align}
 \mathbf{D}^{(v)^\mathrm{T}}\mathbf{D}^{(v)}&=(\mathbf{F}^{(v)}\odot \mathbf{P}^{(v)^\mathrm{T}})(\mathbf{F}^{(v)}\odot \mathbf{P}^{(v)}) \nonumber \\
& = (\mathbf{F}^{(v)^\mathrm{T}}\mathbf{F}^{(v)}) \ast (\mathbf{P}^{(v)^\mathrm{T}} \mathbf{P}^{(v)})
\end{align}
where $\ast$ denotes the Hadamard product or element-wise product of two matrices.

Then the auxiliary matrix $\mathbf{P}^{(v)}$ can be optimized successively in a similar way
\begin{align}
 \mathbf{P}^{(v)}_{t+1}\leftarrow \mathbf{P}^{(v)}_{t}-\frac{1}{L^{(v)}} (2 \mathbf{P}^{(v)}_{t}\mathbf{A}^{(v)}-\mathbf{B}^{(v)})
 \label{P}
\end{align}
where $\mathbf{A}^{(v)} =\mathbf{E}^{(v)^\mathrm{T}} \mathbf{E}^{(v)}+\frac{\mu}{2}(\mathbf{I})$ and $\mathbf{B}^{(v)}=2\mathbf{X}^{(v)}_{(2)}\mathbf{E}^{(v)}+\mu\mathbf{H}^{(v)}+\mathbf{U}^{(v)}$. And $\mathbf{E}^{(v)} =\mathbf{F}^{(v)}\odot \mathbf{H}^{(v)} \in \mathbb{R}^{(NM)\times R}$ and $\mathbf{X}^{(v)}_{(2)} \in \mathbb{R} ^{M \times (MN)}$ is the mode-2 matricization of $\mathcal{X}^{(v)}$.

Moreover, we update the Lagrange multipliers $\mathbf{U}^{(v)}$ using the gradient descent method by
\begin{align}
 \mathbf{U}^{(v)}_t \leftarrow \mathbf{U}^{(v)}_t + \mu(\mathbf{H}^{(v)}-\mathbf{P}^{(v)})
  \label{updateU}
\end{align}

\subsubsection{Fixing $\mathbf{F}^*$ and $\mathbf{H}^{(v)}$, compute $\mathbf{F}^{(v)}$}
By fixing $\mathbf{F^*}$ and $\mathbf{H}^{(v)}$, We minimize the following objective function
\begin{align}
\min_{\mathbf{F}^{(v)}}~||\mathbf{X}^{(v)}_{(3)}-\mathbf{F}^{(v)}\mathbf{J}^{(v)^\mathrm{T}}||_F^2+\lambda_{(v)}||\mathbf{F}^{(v)}-\mathbf{F}^*||_F^2
\end{align}
where $\mathbf{X}^{(v)}_{(3)} \in \mathbb{R}^{N \times (MM)}$ is the mode-3 matricization of tensor $\mathcal{X}^{(v)}$ and $\mathbf{J}^{(v)}=\mathbf{P}^{(v)}\odot\mathbf{H}^{(v)} \in \mathbb{R}^{(MM)\times R}$.

Such an optimization problem can be solved in a similar way as Eq.~(\ref{HE}), from which we get the update rule of $\mathbf{F}^{(v)}$ as follows
\begin{align}
\mathbf{F}_{t+1}^{(v)} \leftarrow \mathbf{F}_{t}^{(v)}-\frac{1}{L^{(v)}}(2 \mathbf{F}_{t}^{(v)}\mathbf{A}^{(v)}-\mathbf{B}^{(v)})
\label{eq:F}
\end{align}
where $\mathbf{A}^{(v)}=\mathbf{J}^{(v)^\mathrm{T}}\mathbf{J}^{(v)}+\lambda_{(v)}(\mathbf{I})$, $\mathbf{B}^{v}=2\mathbf{X}^{(v)}_{(3)}\mathbf{J}^{(v)}+2\lambda_{(v)}\mathbf{F}^{*}$, and $L^{(v)}$ is the maximum eigenvalue of 2$\mathbf{A}^{(v)}$.

\subsubsection{Fixing $\mathbf{H}^{(v)}$ and $\mathbf{F}^{(v)}$, minimize $\mathcal{O}$ over $\mathbf{F}^*$} When $\mathbf{H}^{(v)}$ and $\mathbf{F}^{(v)}$ are fixed, the problem in Eq.~(\ref{eq:obj}) is reduce to a convex optimization problem with respect to $\mathbf{F}^*$. By taking the derivative of the objective function $\mathcal{O}$ in Eq.~(\ref{eq:obj}) with respect to $\mathbf{F}^{*}$ and setting it to zero, we get

\begin{align}
 \mathbf{F^{*}}= \frac{\sum_{v=1}^V \lambda_{(v)}\mathbf{F}^{(v)}}{\sum_{v=1}^{V}\lambda_{(v)}}
 \label{updateF*}
\end{align}

Based on the above analysis, we outline the optimization framework for multi-view multi-graph brain network embedding in Algorithm \ref{algo:admm}. 

\renewcommand{\algorithmicrequire}{\textbf{Input:}}
\renewcommand{\algorithmicensure}{\textbf{Output:}}
\begin{algorithm}[t]
\caption{{\ours}}
\label{algo:admm}
\begin{algorithmic}[1]
\REQUIRE Partically-symmetric tensor $\mathcal{X}^{(v)}$, weight parameters $\alpha_1$ and $\alpha_2$, and embedding dimension $R$
\STATE Initialize $\mathbf{H}^{(v)}, \mathbf{F}^{(v)} \sim\mathcal{N}(0,1), \mathbf{U}=\mathbf{0}, \mu=10$
\REPEAT
    \STATE Update $\mathbf{H}^{(v)}$ and $\mathbf{P}^{(v)}$ by Eq.~(\ref{H}) and Eq.~(\ref{P})
    \STATE Update $\mathbf{U}^{(v)}$ by Eq.~(\ref{updateU})
    \STATE Update $\mathbf{F}^{(v)}$ by Eq.~(\ref{eq:F}) 
    \STATE Update $\mathbf{F}^*$ by Eq.~(\ref{updateF*})
\UNTIL{convergence}
\ENSURE  Consensus embedding matrix $\mathbf{F}^*$ 
\end{algorithmic}
\end{algorithm}

\subsubsection{Computational Analysis}
Algorithm \ref{algo:admm} iteratively solves Eq. (8). In each iteration solving $H^{(v)}$, $P^{(v)}$ and $F^{(v)}$ all requires $O(R^3 + R^2(2M+N+1) + M^2NR)$, solving $U^{(v)}$ requires $O(MR)$, and solving $F^*$ requires $O(NRV)$. Overall, the time complexity involves $O(MaxIter (R^3 + R^2(2M+N+1) + (M^2N + M + NV)R) V)$, which is linear to the number of nodes $N$, so the proposed method is applicable for larger scale of brain network. 

\section{Experiments and Evaluation} \label{sec:exp}
In order to empirically evaluate the performance of the proposed {\ours} approach for multi-view multi-graph brain network clustering analysis, we test our model on two real datasets, HIV and Bipolar disorder with fMRI brain networks and DTI brain networks, and compare with several state-of-the-art multi-view clustering methods.
\begin{table*}[htbp]
  \centering
  \caption{Clustering Accuracy and F1 score on HIV dataset and BP dataset}
   \resizebox{7in}{!}{
    \begin{tabular}{r|l|cccccccccc}
    \hline
        &      & \multicolumn{1}{|l}{Method} &       &       &       &       &       &       &       &  \\
        \cline{3-12}
    \multicolumn{1}{c}{Dataset} & \multicolumn{1}{|c}{Measure} & \multicolumn{1}{|c}{SEC} & \multicolumn{1}{c}{convexSub} & \multicolumn{1}{c}{AMGL} & \multicolumn{1}{c}{multiNMF} & \multicolumn{1}{c}{CoRegSc} & \multicolumn{1}{c}{MIC} & \multicolumn{1}{c}{SCMV-3DT} & \multicolumn{1}{c}{M2E-TS}&\multicolumn{1}{c}{M2E-DS} & \multicolumn{1}{c}{M2E} \\
    \hline
          & Accuracy  & 50.00(8) & 52.86(7) & 52.86(7) & 57.23(4) & 57.14(5) & 55.72(6) & 64.29(3) & 52.86(7) & 68.57(2) & \textbf{71.43(1)} \\
    & F1   & 49.74(8) & 53.52(7) & 53.52(7) & 56.18(6) & 60.53(4) & 58.28(5) & 66.67(3) &49.23(9) & 68.57(2) & \textbf{72.22(1)} \\
    & Precision  & 50.00(9) & 52.77(7) &52.77(7)  & 57.37(5) & 56.10(5) & 61.12(4) & 62.50(3) & 53.25(6) & 68.57(2) & \textbf{69.73(1)} \\
     \multicolumn{1}{l|}{HIV} & Recall  & 49.86(8) & 54.29(7) & 54.29(7) & 55.98(5) & 65.71(4) & 56.28(6) & 71.43(2) & 45.57(9) & 68.57(3) &\textbf{75.00(1)} \\
          \hline
          & Accuracy &  54.64(7) & 52.57(9) & 52.77(8) & 58.52(4) & 56.70(6) & 61.86(2) & 54.64(7) & 57.73(5) & 60.82(3) & \textbf{68.04(1)} \\
   & F1   & 56.86(7) & 28.46(8) & 28.13(9) & 66.99(3) & 60.38(6) & \textbf{72.59(1)} & 59.26(6) & 64.34(4) & 61.22(5)  & 68.69(2) \\
            & Precision  & 58.00(9) & \textbf{74.78(1)} & 74.83(2) & 58.34(8) & 59.25(5) & 59.03(6) & 57.14(9) & 58.73(7)  & 65.22(4) &72.34(3)\\
      \multicolumn{1}{l|}{BP} & Recall  &55.76(7)  &17.84(8)  & 17.71(9) & 80.40(2) & 61.53(5) & \textbf{94.23(1)} & 61.53(5) & 71.15(3) & 57.69(6) & 65.38(4)\\
          \hline
    \end{tabular}%
    }
  \label{tab:compare_r}%
\end{table*}

\subsection{Data Collection and Preprocessing}
\begin{itemize}[leftmargin=*]
\item \textbf{Human Immunodeficiency Virus Infection (HIV)}: The original dataset is unbalanced, we randomly sampled 35 patients and 35 controls from the dataset for performance evaluation. A detailed description about data acquisition is available in \cite{cao2015identifying}. For fMRI data, we used DPARSF \footnote{\url{http://rfmri.org/DPARSF}} and SPM \footnote{\url{http://www.l.ion.ucl.ac.uk/spm/software/spm8}} toolboxes for preprocessing. We construct each graph with $90$ nodes where links are created based on the correlations between different brain regions. For DTI data, we used FSL toolbox\footnote{\url{https://fsl.fmrib.ox.ac.uk/fsl/fslwiki/}} for preprocessing and parcellated each DTI image into $90$ regions by the AAL \cite{tzourio2002automated}.
 
\item  \textbf{Bipolar Disorder (BP)}: This dataset consists of 52 bipolar subjects who are currently in euthymia and 45 healthy controls. For fMRI data, we used the toolbox CONN \footnote{\url{http://www.nitrc.org/projects/conn}} to construct fMRI data of the BP brain network \cite{whitfield2012}. Using the $82$ labels Freesurfer-generated cortical/subcortical gray matter regions, functional brain networks were derived using pairwise BOLD signal correlations. For DTI, same as fMRI, we constructed the DTI image into $82$ regions.
\end{itemize}

\subsection{Baselines and Metrics}
We compare the proposed {\ours} with eight other methods for multi-view clustering on brain networks. We adopt accuracy and F1-score as our evaluation metrics.

$(1)$ \textbf{SEC} is a single-view spectral embedding clustering framework \cite{nie2011spectral}. $(2)$ \textbf{convexSub} is a convex subspace representation learning method \cite{guo2013convex}. $(3)$ \textbf{CoRegSc} is a co-regularization based multi-view spectral clustering framework \cite{kumar2011co}. $(4)$ \textbf{MultiNMF} is the NMF-based multi-view clustering method by searching for a factorization that gives compatible clustering solutions across multiple views \cite{liu2013multi}. $(5)$ \textbf{MIC} first uses the kernel matrices to form an initial tensor across all the multiple sources \cite{shao2015clustering}. $(6)$ \textbf{AMGL} is a recently proposed non-parameter multi-view spectral learning framework \cite{AMGL}. $(7)$ \textbf{SCMV-3DT} uses t-product in the third-order tensor space and represents multi-view data by a t-linear combination with sparse and low-rank \cite{yin2016low}. $(8)$ \textbf{M2E-TS} is the two-step method of {\ours}. $(9)$ \textbf{M2E-DS} is the directly shared method of {\ours} mentioned on Section Methodology. 

Since SEC is designed for single-view data, we first concatenate all the views together and then apply SEC on the concatenated views. For all the spectral clustering based methods, we construct the RBF kernel matrices with kernel width $\sigma$ to be the median distance among all the brain network samples. Following \cite{von2007tutorial}, we construct each graph by selecting $10$-nearest neighbors among raw data. We tune the parameters of each baseline methods using the strategy mentioned in the corresponding paper. There are three main parameters in our model, namely $\lambda_1$, $\lambda_2$ and $R$, where $\lambda_1$ and $\lambda_2$ are the weight parameters reflecting the importance of different views, and $R$ is the number of factors representing the embedded dimension. We apply the grid search to determine the optimal values of these three parameters. In particular, we empirically select $\lambda_1$ and $\lambda_2$ from $\{10^{-4},10^{-2},...,10^{4}\}$, and $R$ is selected from $\{1,2,...,20\}$. For evaluation, since there are two possible label values, normal and control, for each brain network sample on both HIV and BP datasets, we set the number of clusters $K$ to be $2$ and test how well our method can group the brain networks of patients and normal controls into two different clusters.

In order to make a fair comparison, we apply the ``Litekmeans'' function in Matlab \cite{cai2011litekmeans} for all the compared methods during their $K$-means clustering step. We repeat this $K$-means clustering procedure 20 times with random initialization, as ``Litekmeans'' greatly depends on initialization. For the evaluation, we repeat running the program of each clustering methods 20 times and report the average Accuracy, F1 score, Precision and Recall as the results. 

\subsection{Clustering Results}
Table {\ref{tab:compare_r}} shows the clustering results. We see that in terms of accuracy, the proposed {\ours} method performs better than all the other baseline methods on both HIV and BP datasets.  The single-view clustering SEC does not distinguish the features from different views, which leads to a poor performance than the multi-view methods. Compared with subgraph method convexSub, {\ours} achieves better performance, which may because our method can capture the complex multi-way relation in brain networks. The common property of three multi-view clustering methods, AMGL, multiNMF and CoRegSc, is that the features they learned for each view are based on vector representations. However, for graph instances, the structural information is hardly persevered by the flattened vector representations, which could be the underlying reason that these three methods cannot outperform {\ours}. Moreover, by using tensor techniques to model the multi-view multi-graph collectively, {\ours} could learn discriminative latent representations and graph-specific features. While MIC and SCMV-3DT also use tensor to represent the multi-view learning, their performance are not beyond {\ours}. This is mainly because they still learn the vector representation of each graph, which makes them fail to capture the multi-graph structure. M2E-TS cannot explore multiple views and multiple graphs simultaneously; therefore it gets a worse clustering result than {\ours}. Besides, by comparing with M2E-DS, the proposed {\ours} utilizing the constraint to regularize clustering solutions obtained from multiple views towards a consensus solution can find the true clustering more effectively than simply concatenating all the features together.

Note that in terms of F1 score, {\ours} is the best one on HIV dataset, while MIC is the best on BP dataset. This is because MIC method clusters large number of samples into the same class while M2E produces relatively even results. Besides, the precision-recall result also explains why our M2E does not outperform MIC in terms of F1 score. And the other methods like convexSub, AMGL and multiNMF have the same problem as MIC, therefore their precision or recall results are higher than M2E. However, in our context, it’s not reasonable to cluster all samples in one class, in that it cannot distinguish patients from controls.

\subsection{Parameter Sensitivity Analysis}
In this section, to evaluate how the parameters of {\ours} affects performance, we study the sensitivity of the three main parameters, including $\lambda_1$, $\lambda_2$ and $R$, where $\lambda_1$ is the parameter of DTI view, $\lambda_2$ is the parameter of fMRI view and $R$ is the embedded dimension. 
For evaluating the regularization parameter $\lambda_1$ and $\lambda_2$, we set the $R$ to the optimal value.

\begin{figure}[t]
    \centering
    \begin{subfigure}[t]{.4\columnwidth} 
        \centering
        \includegraphics[width=\textwidth]{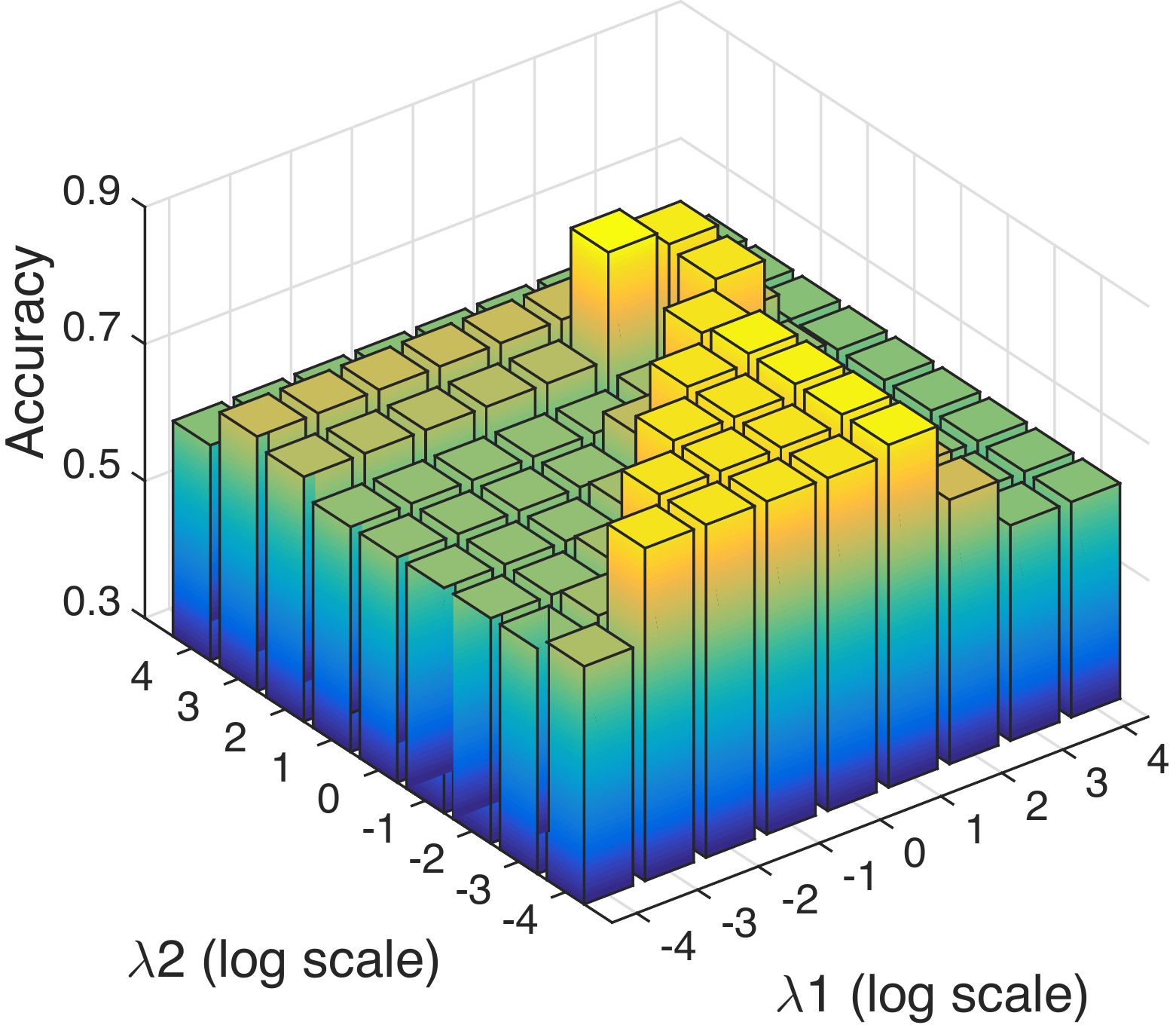} 
        \caption{HIV} \label{fig:HIV_alpha}
    \end{subfigure}
    ~ 
    \begin{subfigure}[t]{.4\columnwidth}
        \centering
        \includegraphics[width=\textwidth]{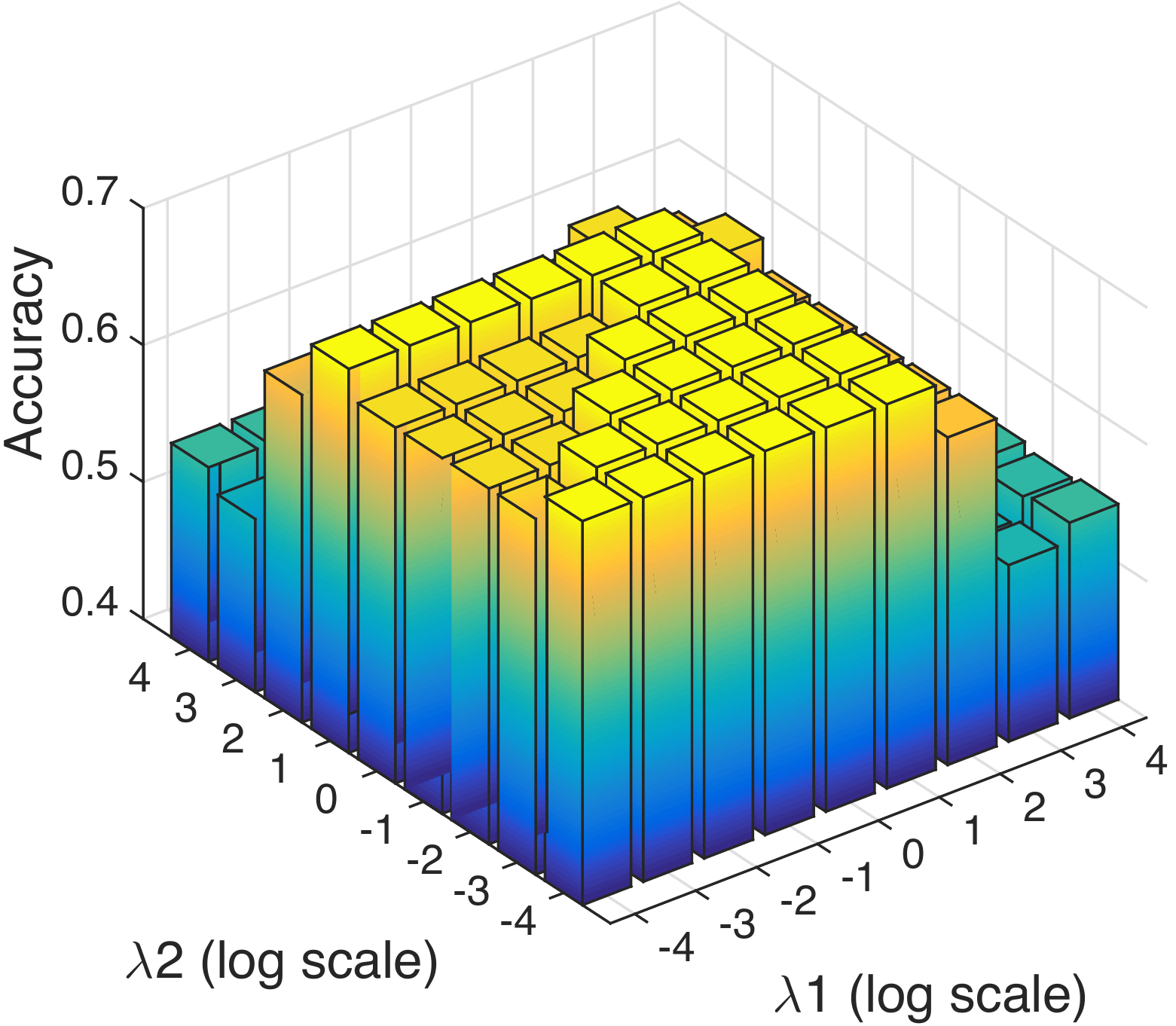}
        \caption{BP}  \label{fig:BP_alpha}
    \end{subfigure}
    \caption{Accuracy with different weights $\lambda_1$ and $\lambda_2$ on HIV dataset and BP dataset}
\end{figure}

\begin{figure}[t]
    \centering
    \begin{subfigure}[t]{.4\columnwidth} 
        \centering
        \includegraphics[width=\textwidth]{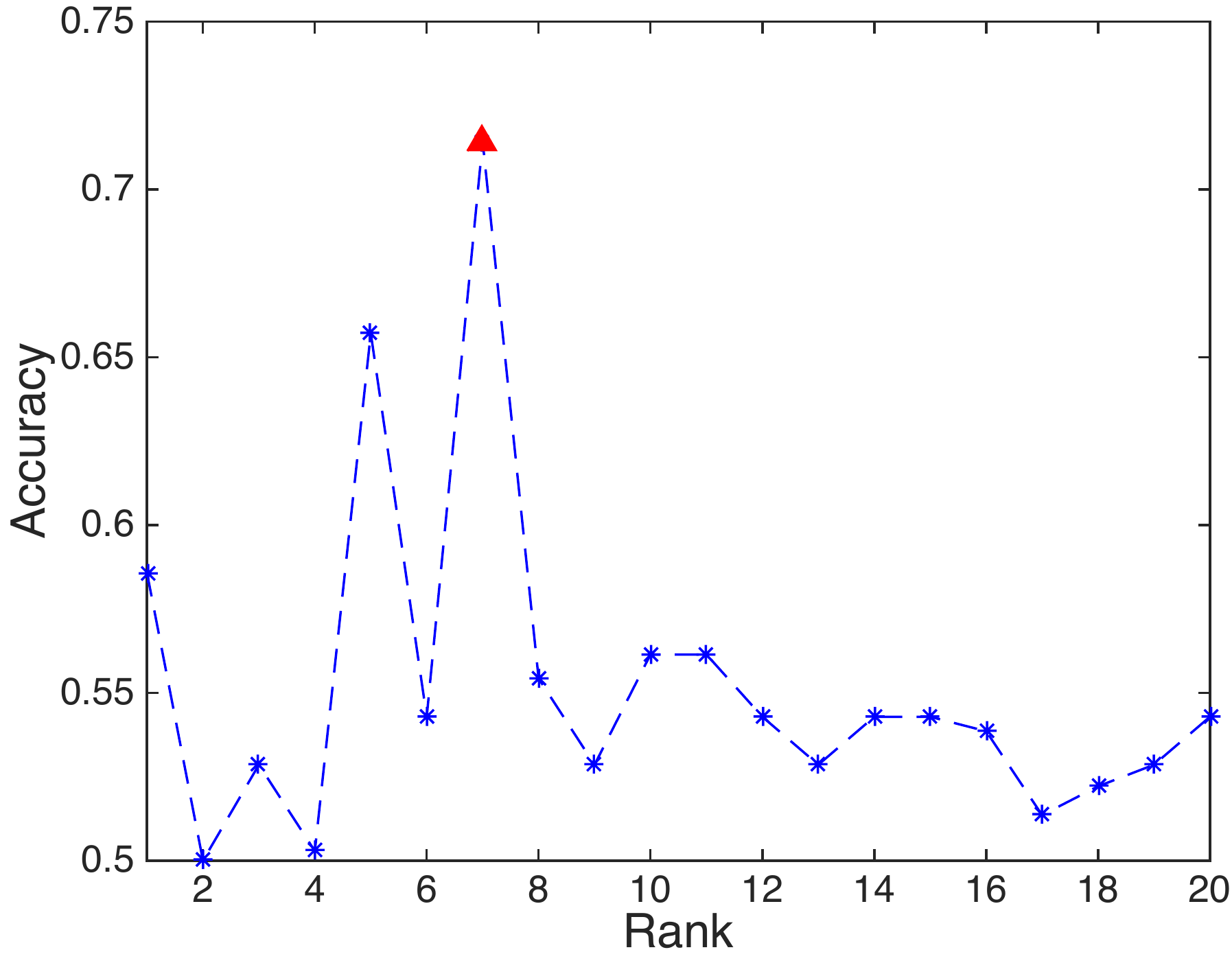}
        \caption{HIV} \label{fig:HIV_rank}
    \end{subfigure}
    ~ ~ ~
    \begin{subfigure}[t]{.4\columnwidth} 
        \centering
        \includegraphics[width=\textwidth]{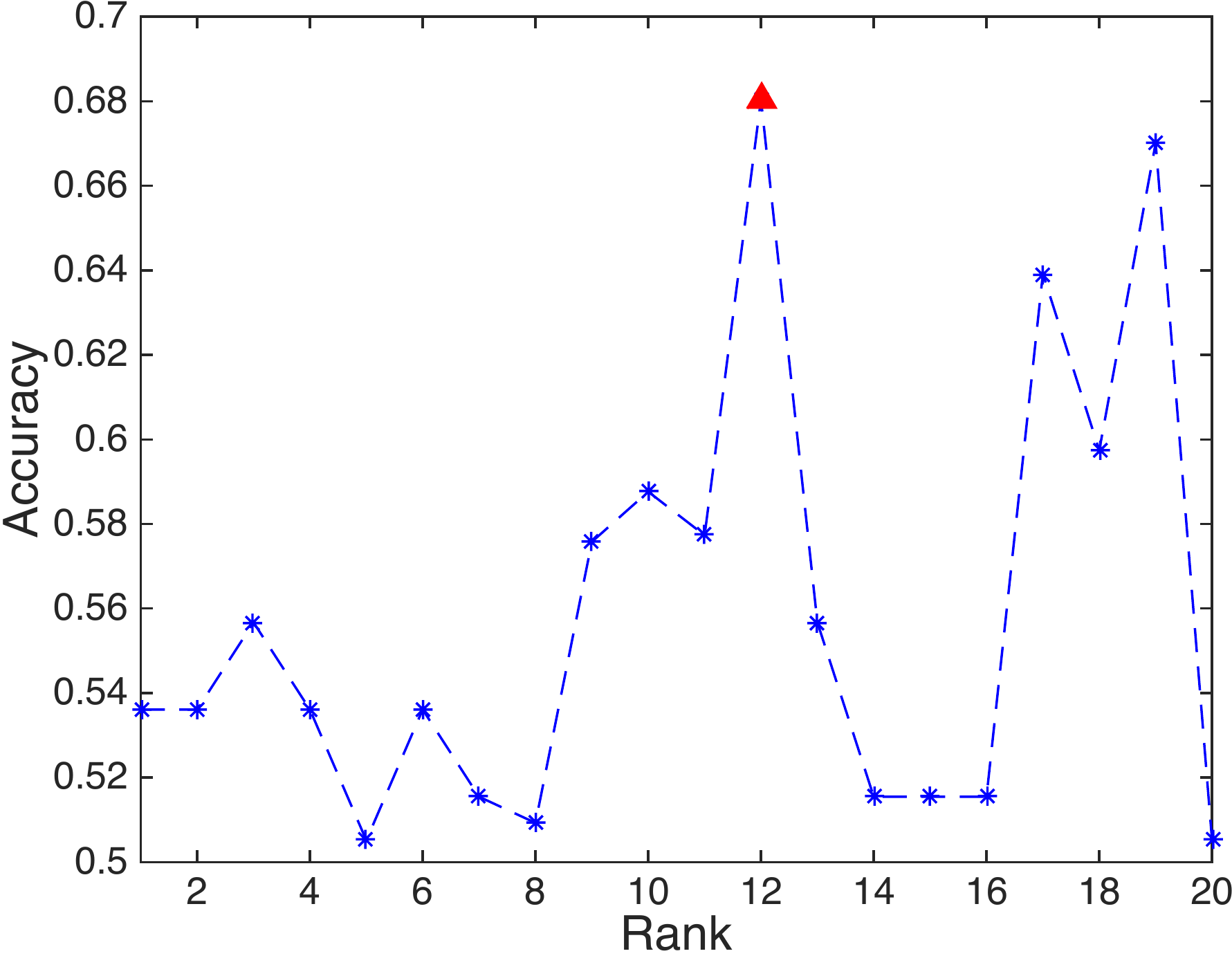}
        \caption{BP}\label{fig:BP_rank}
    \end{subfigure}
    \caption{Accuracy with different embedded dimensions $R$ on HIV dataset and BP dataset}
\end{figure}

According to Figures \ref{fig:HIV_alpha} and \ref{fig:BP_alpha}, we observe that the result is relatively sensitive to the change of $\lambda_1$ and $\lambda_2$, which shows that different views have different effects on the performance. Besides, when $\lambda_1$ and $\lambda_2$ are very large, the performance gets worse. This shows that $\lambda_1$ and $\lambda_2$ are also important for tuning the first term and the second term of objective function in Eq.~(\ref{eq:obj}).

Figures \ref{fig:HIV_rank} and \ref{fig:BP_rank} show the performance of {\ours} with the $R$ value varying from $1$ to $20$. We can observe that the embedded dimension has a significant effect on the accuracy. The highest accuracy is achieved when $R$ equals to $7$ on HIV dataset and $12$ on BP dataset. Generally speaking, the performance shakes greatly with the change of the rank. But in most cases the optimal value of $R$ lies in a small range of values as demonstrated in \cite{hao2013linear} and it is not time-consuming to find it using the grid search strategy in practical applications. 

\subsection{Factor Analysis}
\begin{figure}[t]
    \centering
    \begin{subfigure}[t]{.6\columnwidth} 
        \centering
        \includegraphics[width=\textwidth]{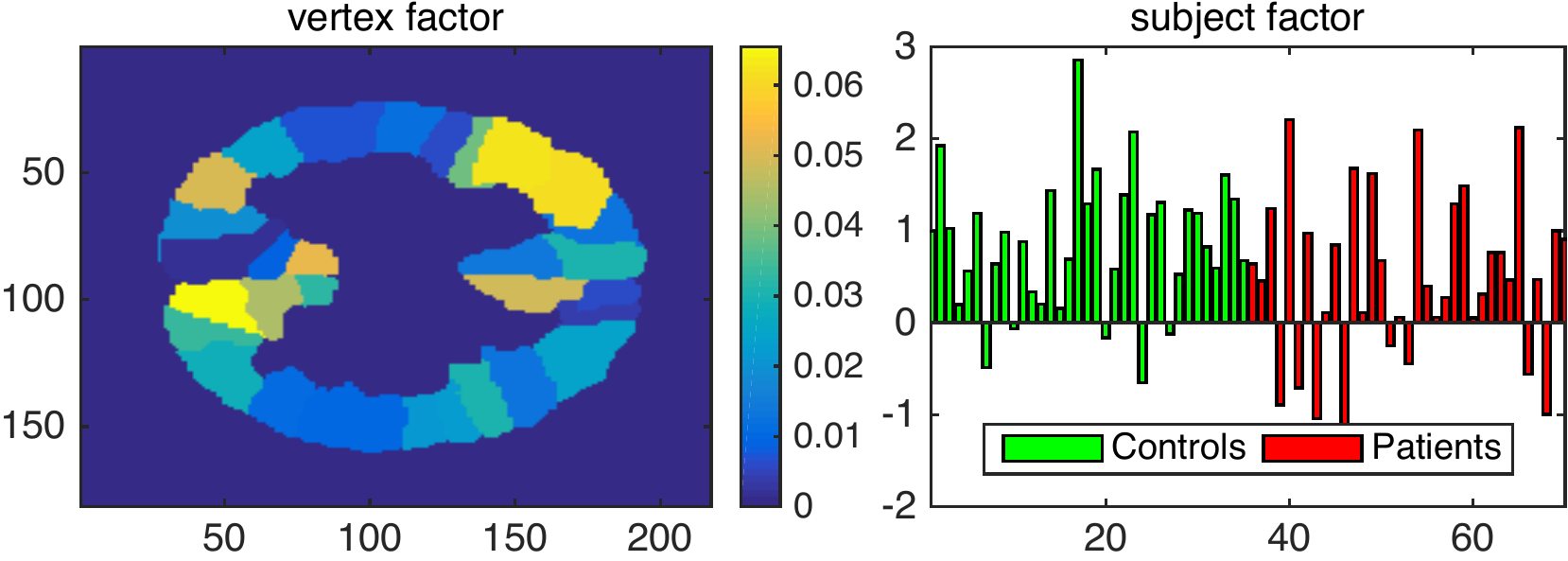}
        \caption{fMRI}\label{fig:HIV_fMRI_factor}
    \end{subfigure}
    ~ 
    \begin{subfigure}[t]{.6\columnwidth} 
        \centering
        \includegraphics[width=\textwidth]{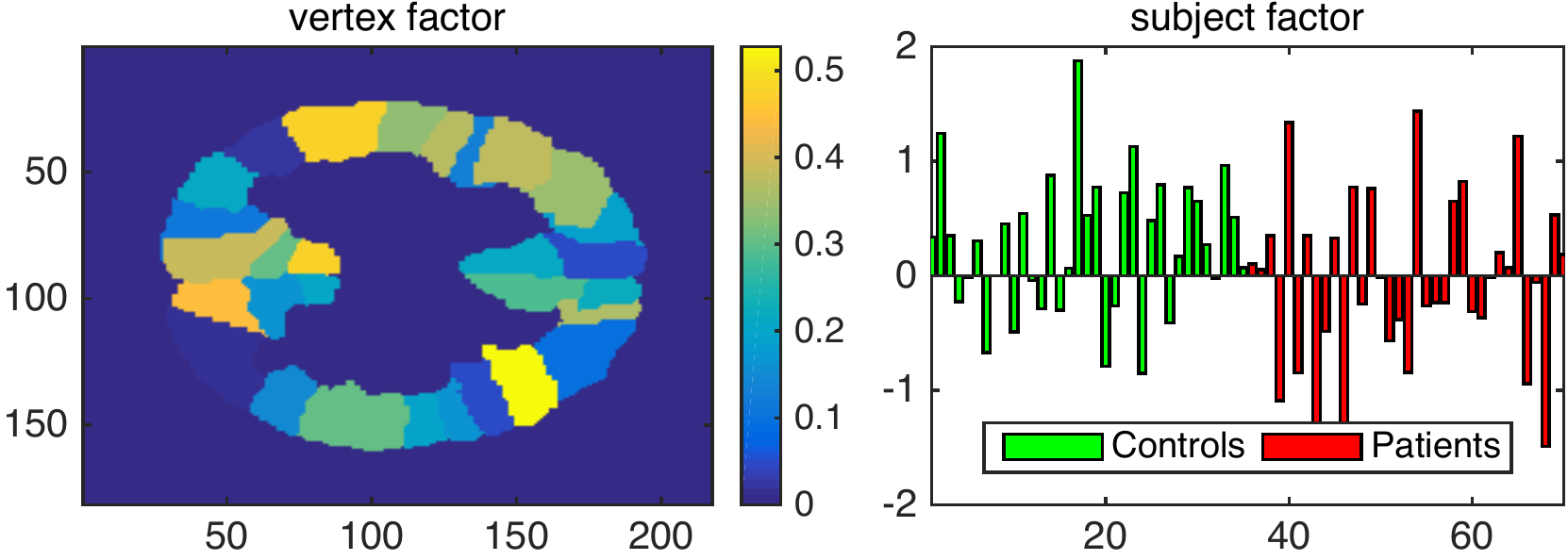}
        \caption{DTI}\label{fig:HIV_DTI_factor}
    \end{subfigure}
    \caption{Embedded features of nodes learning $\mathbf{H}$ and graphs learning $\mathbf{F}$ (left and right panels respectively) from fMRI and DTI on HIV dataset}
\end{figure}

{\ours} extracts $\mathbf{H}^{(v)}$ consisting of $\mathbf{h}_r^{(v)}$, for $r= 1,...,R$ and $\mathbf{F}^{(v)}$ consisting of $\mathbf{f}_r^{(v)}$, where these factors indicate the signatures of sources in vertex and subject domain, respectively. Due to limited space, we only show the factors on HIV dataset here. We visualize the learning results of $\mathbf{H}^{(v)}$ and $\mathbf{F}^{(v)}$ on fMRI and DTI dataset in Figures \ref{fig:HIV_fMRI_factor} and \ref{fig:HIV_DTI_factor}.

We show the largest factors in terms of magnitude for fMRI and DTI in Figures \ref{fig:HIV_fMRI_factor} and \ref{fig:HIV_DTI_factor}. Left panel shows the node embedded feature $\mathbf{H}^{(v)}$. The coordinate system represents neuroanatomy and the color shows the activity intensity of the brain region. The right panel shows the graph embedded feature $\mathbf{F}^{(v)}$ which represents the factor strengths for both patients and controls. Based on our objective function in Eq.~(\ref{eq:obj}), $\mathbf{H}^{(v)}$ is used to preserve the individual information of each view and dependencies among multiple graphs. As we can see from the left panels of Figures \ref{fig:HIV_fMRI_factor} and \ref{fig:HIV_DTI_factor}, the embedded neuroanatomy learned from fMRI data and DTI data are widely different from each other. However, $\mathbf{F}^{(v)}$ is learned by forcing each view to the consensus correlation $\mathbf{F}^*$. From the right panels of Figures \ref{fig:HIV_fMRI_factor} and \ref{fig:HIV_DTI_factor}, results from both views show that the controls have relatively positive correlation with node embedded feature, while the patient have relatively negative correlation. Moreover, those neuroimaging findings in HIV generally support clinical observations of functional impairments in attention, psychomotor speed, memory, and executive function. In particular, regions identified in our current study are consistent with those reported in structural and functional MRI studies of HIV associated neurocognitive disorder (HAND), including regions within the frontal and parietal lobes \cite{risacher2013neuroimaging}.

\section{Conclusion} \label{sec:con}
We present a novel multi-view multi-graph embedding framework based on partially-symmetric tensor factorization for brain network analysis. The proposed {\ours} method not only takes advantages of the complementary and dependent information among multiple views and multiple graphs, but also exploits the graph structures. In particular, we first model the multi-view multi-graph data as multiple partially-symmetric tensors, and then learn the consensus graph embedding via the integration of tensor factorization and a multi-view embedding method. We apply our approach on two real HIV and BP datasets with a fMRI view and a DTI view for unsupervised brain network analysis. Extensive experimental results demonstrate the effectiveness of {\ours} for multi-view multi-graph embedding on brain networks.

\section*{Acknowledgements}
This work is supported in part by NSF grants No. IIS-1526499 and CNS-1626432, NIH grant No. R01-MH080636, and NSFC grants No. 61503253 and 61672313.

\bibliographystyle{aaai}
\bibliography{Liu-He}

\begin{thebibliography}{}

\bibitem[\protect\citeauthoryear{Boyd \bgroup et al\mbox.\egroup
  }{2011}]{boyd2011distributed}
Boyd, S.; Parikh, N.; Chu, E.; Peleato, B.; and Eckstein, J.
\newblock 2011.
\newblock Distributed optimization and statistical learning via the alternating
  direction method of multipliers.
\newblock {\em Foundations and Trends{\textregistered} in Machine Learning}
  3(1):1--122.

\bibitem[\protect\citeauthoryear{Cai}{2011}]{cai2011litekmeans}
Cai, D.
\newblock 2011.
\newblock Litekmeans: the fastest matlab implementation of kmeans.
\newblock {\em Software available at: http://www. zjucadcg.
  cn/dengcai/Data/Clustering. html}.

\bibitem[\protect\citeauthoryear{Cao \bgroup et al\mbox.\egroup
  }{2014}]{cao2014tensor}
Cao, B.; He, L.; Kong, X.; Philip, S.~Y.; Hao, Z.; and Ragin, A.~B.
\newblock 2014.
\newblock Tensor-based multi-view feature selection with applications to brain
  diseases.
\newblock In {\em ICDM},  40--49.
\newblock IEEE.

\bibitem[\protect\citeauthoryear{Cao \bgroup et al\mbox.\egroup
  }{2015}]{cao2015identifying}
Cao, B.; Kong, X.; Zhang, J.; Yu, P.~S.; and Ragin, A.~B.
\newblock 2015.
\newblock Identifying hiv-induced subgraph patterns in brain networks with side
  information.
\newblock {\em Brain informatics} 2(4):211--223.

\bibitem[\protect\citeauthoryear{Guo}{2013}]{guo2013convex}
Guo, Y.
\newblock 2013.
\newblock Convex subspace representation learning from multi-view data.
\newblock In {\em AAAI}, volume~1, ~2.

\bibitem[\protect\citeauthoryear{Hao \bgroup et al\mbox.\egroup
  }{2013}]{hao2013linear}
Hao, Z.; He, L.; Chen, B.; and Yang, X.
\newblock 2013.
\newblock A linear support higher-order tensor machine for classification.
\newblock {\em IEEE Transactions on Image Processing} 22(7):2911--2920.

\bibitem[\protect\citeauthoryear{Hartigan and
  Wong}{1979}]{hartigan1979algorithm}
Hartigan, J.~A., and Wong, M.~A.
\newblock 1979.
\newblock Algorithm as 136: A k-means clustering algorithm.
\newblock {\em Journal of the Royal Statistical Society. Series C (Applied
  Statistics)} 28(1):100--108.

\bibitem[\protect\citeauthoryear{He \bgroup et al\mbox.\egroup
  }{2014}]{he2014dusk}
He, L.; Kong, X.; Yu, P.~S.; Yang, X.; Ragin, A.~B.; and Hao, Z.
\newblock 2014.
\newblock Dusk: A dual structure-preserving kernel for supervised tensor
  learning with applications to neuroimages.
\newblock In {\em SDM},  127--135.
\newblock SIAM.

\bibitem[\protect\citeauthoryear{He \bgroup et al\mbox.\egroup
  }{2017}]{he2017kernelized}
He, L.; Lu, C.-T.; Ma, G.; Wang, S.; Shen, L.; Philip, S.~Y.; and Ragin, A.~B.
\newblock 2017.
\newblock Kernelized support tensor machines.
\newblock In {\em ICML},  1442--1451.

\bibitem[\protect\citeauthoryear{Jie \bgroup et al\mbox.\egroup
  }{2014}]{jie2014integration}
Jie, B.; Zhang, D.; Gao, W.; Wang, Q.; Wee, C.-Y.; and Shen, D.
\newblock 2014.
\newblock Integration of network topological and connectivity properties for
  neuroimaging classification.
\newblock {\em IEEE Transactions on Biomedical Engineering} 61(2):576--589.

\bibitem[\protect\citeauthoryear{Kong and Yu}{2014}]{kong2014brain}
Kong, X., and Yu, P.~S.
\newblock 2014.
\newblock Brain network analysis: a data mining perspective.
\newblock {\em ACM SIGKDD Explorations Newsletter} 15(2):30--38.

\bibitem[\protect\citeauthoryear{Kumar and Daum{\'e}}{2011}]{kumar2011traing}
Kumar, A., and Daum{\'e}, H.
\newblock 2011.
\newblock A co-training approach for multi-view spectral clustering.
\newblock In {\em ICML},  393--400.

\bibitem[\protect\citeauthoryear{Kumar, Rai, and Daume}{2011}]{kumar2011co}
Kumar, A.; Rai, P.; and Daume, H.
\newblock 2011.
\newblock Co-regularized multi-view spectral clustering.
\newblock In {\em NIPS},  1413--1421.

\bibitem[\protect\citeauthoryear{Kuo \bgroup et al\mbox.\egroup
  }{2015}]{kuo2015unified}
Kuo, C.-T.; Wang, X.; Walker, P.; Carmichael, O.; Ye, J.; and Davidson, I.
\newblock 2015.
\newblock Unified and contrasting cuts in multiple graphs: application to
  medical imaging segmentation.
\newblock In {\em KDD},  617--626.
\newblock ACM.

\bibitem[\protect\citeauthoryear{Lin, Liu, and Su}{2011}]{lin2011linearized}
Lin, Z.; Liu, R.; and Su, Z.
\newblock 2011.
\newblock Linearized alternating direction method with adaptive penalty for
  low-rank representation.
\newblock In {\em NIPS},  612--620.

\bibitem[\protect\citeauthoryear{Liu \bgroup et al\mbox.\egroup
  }{2013a}]{liu2013multi}
Liu, J.; Wang, C.; Gao, J.; and Han, J.
\newblock 2013a.
\newblock Multi-view clustering via joint nonnegative matrix factorization.
\newblock In {\em SDM},  252--260.
\newblock SIAM.

\bibitem[\protect\citeauthoryear{Liu \bgroup et al\mbox.\egroup
  }{2013b}]{liu2013mining}
Liu, W.; Chan, J.; Bailey, J.; Leckie, C.; and Ramamohanarao, K.
\newblock 2013b.
\newblock Mining labelled tensors by discovering both their common and
  discriminative subspaces.
\newblock In {\em SDM},  614--622.
\newblock SIAM.

\bibitem[\protect\citeauthoryear{Ma \bgroup et al\mbox.\egroup
  }{2016}]{ma2016multi}
Ma, G.; He, L.; Cao, B.; Zhang, J.; Yu, P.~S.; and Ragin, A.~B.
\newblock 2016.
\newblock Multi-graph clustering based on interior-node topology with
  applications to brain networks.
\newblock In {\em ECML-PKDD},  476–492.
\newblock Springer.

\bibitem[\protect\citeauthoryear{Ma \bgroup et al\mbox.\egroup
  }{2017a}]{ma2017multi}
Ma, G.; He, L.; Lu, C.-T.; Shao, W.; Yu, P.~S.; Leow, A.~D.; and Ragin, A.~B.
\newblock 2017a.
\newblock Multi-view clustering with graph embedding for connectome analysis.
\newblock In {\em CIKM}.

\bibitem[\protect\citeauthoryear{Ma \bgroup et al\mbox.\egroup
  }{2017b}]{ma2017multi2}
Ma, G.; Lu, C.-T.; He, L.; Yu, P.~S.; and Ragin, A.~B.
\newblock 2017b.
\newblock Multi-view graph embedding with hub detection for brain network
  analysis.
\newblock In {\em ICDM}.

\bibitem[\protect\citeauthoryear{Mousazadeh and
  Cohen}{2015}]{mousazadeh2015embedding}
Mousazadeh, S., and Cohen, I.
\newblock 2015.
\newblock Embedding and function extension on directed graph.
\newblock {\em Signal Processing} 111:137--149.

\bibitem[\protect\citeauthoryear{Nie \bgroup et al\mbox.\egroup
  }{2011}]{nie2011spectral}
Nie, F.; Zeng, Z.; Tsang, I.~W.; Xu, D.; and Zhang, C.
\newblock 2011.
\newblock Spectral embedded clustering: A framework for in-sample and
  out-of-sample spectral clustering.
\newblock {\em IEEE Transactions on Neural Networks} 22(11):1796--1808.

\bibitem[\protect\citeauthoryear{Nie, Li, and Li}{2016}]{AMGL}
Nie, F.; Li, J.; and Li, X.
\newblock 2016.
\newblock Parameter-free auto-weighted multiple graph learning: A framework for
  multiview clustering and semi-supervised classification.
\newblock In {\em IJCAI}.

\bibitem[\protect\citeauthoryear{Ou \bgroup et al\mbox.\egroup
  }{2016}]{ou2016asymmetric}
Ou, M.; Cui, P.; Pei, J.; Zhang, Z.; and Zhu, W.
\newblock 2016.
\newblock Asymmetric transitivity preserving graph embedding.
\newblock In {\em KDD},  1105--1114.

\bibitem[\protect\citeauthoryear{Parikh, Boyd, and
  others}{2014}]{parikh2014proximal}
Parikh, N.; Boyd, S.; et~al.
\newblock 2014.
\newblock Proximal algorithms.
\newblock {\em Foundations and Trends{\textregistered} in Optimization}
  1(3):127--239.

\bibitem[\protect\citeauthoryear{Risacher and
  Saykin}{2013}]{risacher2013neuroimaging}
Risacher, S.~L., and Saykin, A.~J.
\newblock 2013.
\newblock Neuroimaging biomarkers of neurodegenerative diseases and dementia.
\newblock In {\em Seminars in neurology}, volume~33,  386--416.
\newblock Thieme Medical Publishers.

\bibitem[\protect\citeauthoryear{Shao, He, and Yu}{2015}]{shao2015clustering}
Shao, W.; He, L.; and Yu, P.~S.
\newblock 2015.
\newblock Clustering on multi-source incomplete data via tensor modeling and
  factorization.
\newblock In {\em PAKDD},  485--497.
\newblock Springer.

\bibitem[\protect\citeauthoryear{Sun \bgroup et al\mbox.\egroup
  }{2017}]{sun2017sequential}
Sun, L.; Wang, Y.; Cao, B.; Yu, P.~S.; Srisa-an, W.; and Leow, A.~D.
\newblock 2017.
\newblock Sequential keystroke behavioral biometrics for mobile user
  identification via multi-view deep learning.
\newblock In {\em ECML-PKDD}.

\bibitem[\protect\citeauthoryear{Tzourio-Mazoyer \bgroup et al\mbox.\egroup
  }{2002}]{tzourio2002automated}
Tzourio-Mazoyer, N.; Landeau, B.; Papathanassiou, D.; Crivello, F.; Etard, O.;
  Delcroix, N.; Mazoyer, B.; and Joliot, M.
\newblock 2002.
\newblock Automated anatomical labeling of activations in spm using a
  macroscopic anatomical parcellation of the mni mri single-subject brain.
\newblock {\em Neuroimage} 15(1):273--289.

\bibitem[\protect\citeauthoryear{Van~Loan}{2016}]{van2016structured}
Van~Loan, C.~F.
\newblock 2016.
\newblock Structured matrix problems from tensors.
\newblock In {\em Exploiting Hidden Structure in Matrix Computations:
  Algorithms and Applications}. Springer.
\newblock  1--63.

\bibitem[\protect\citeauthoryear{Von~Luxburg}{2007}]{von2007tutorial}
Von~Luxburg, U.
\newblock 2007.
\newblock A tutorial on spectral clustering.
\newblock {\em Statistics and computing} 17(4):395--416.

\bibitem[\protect\citeauthoryear{Whitfield-Gabrieli and
  Nieto-Castanon}{2012}]{whitfield2012}
Whitfield-Gabrieli, S., and Nieto-Castanon, A.
\newblock 2012.
\newblock Conn: a functional connectivity toolbox for correlated and
  anticorrelated brain networks.
\newblock {\em Brain connectivity} 2(3):125--141.

\bibitem[\protect\citeauthoryear{Yin \bgroup et al\mbox.\egroup
  }{2016}]{yin2016low}
Yin, M.; Gao, J.; Xie, S.; and Guo, Y.
\newblock 2016.
\newblock Low-rank multi-view clustering in third-order tensor space.
\newblock {\em arXiv preprint arXiv:1608.08336}.

\end{thebibliography}
\end{document}